\documentclass[lettersize,journal]{IEEEtran}
\usepackage{amsmath,amsfonts,amssymb}

\usepackage{graphicx}
\usepackage{algorithm}
\usepackage{algorithmic}
\usepackage{siunitx}
\usepackage[caption=false,font=normalsize,labelfont=sf,textfont=sf]{subfig}
\usepackage{textcomp}
\usepackage{stfloats}
\usepackage{url}

\usepackage[hidelinks]{hyperref}
\usepackage{booktabs}
\usepackage{multirow}
\usepackage{array}

\usepackage{cite}
\usepackage{tabularx}

\begin{document}
\title{ColorFD: A Finite-Difference Guided Black-Box Physical Adversarial Attack for Remote Sensing Object Detection}

\author{Tiannuo~Guo,
        Guhang~Qiu,
        Yuzhen~Xie,
        Rui~Feng,
        Ligang~Li,
        and~Deliang~Xiang,~\IEEEmembership{Member,~IEEE}%
\thanks{The authors are with the College of Information Science and Technology, Beijing University of Chemical Technology, Beijing 100029, China (e-mail: mingzhixin2002@gmail.com; 2024200847@buct.edu.cn; 2024400282@buct.edu.cn; 2025400282@buct.edu.cn; liligang@buct.edu.cn; xiangdeliang@buct.edu.cn).}%
\thanks{Corresponding author: Deliang Xiang (e-mail: xiangdeliang@buct.edu.cn).}%
\thanks{This work was supported by the National Natural Science Foundation of China under Grant No. 62171015.}}

\markboth{IEEE Transactions on Circuits and Systems for Video Technology}%
{Guo \MakeLowercase{\textit{et al.}}: ColorFD}

\maketitle

\begin{abstract}
	Although deep neural network-based remote sensing object detectors have achieved strong performance, they remain vulnerable to adversarial perturbations. Existing studies mainly focus on digital or white-box settings, whereas black-box physical attacks remain underexplored. These attacks are often constrained by limited physical feasibility and inefficient optimization in high-dimensional search spaces. To address these challenges, this paper proposes ColorFD, a black-box physical attack based on multiple pure-color patches. The patch positions and color parameters are jointly optimized using Differential Evolution (DE). A target-wise fitness and selection mechanism evaluates the attack state of each target and preserves target-specific improvements during evolution. Two guidance strategies further constrain the patch search space. Key-region localization identifies sensitive regions through finite-difference color probing. Common-feature extraction provides category-level spatial priors and avoids repeated localization. Although evaluated on aircraft, the formulation is not inherently restricted to this category. Experiments on YOLOv3u, YOLOv5u, and Faster R-CNN show that ColorFD outperforms the tested black-box patch method across all evaluated detectors and remains competitive with strong white-box baselines. Physical-world experiments further demonstrate that the optimized pure-color patches can be transferred from the digital domain to real imaging conditions.
\end{abstract}


\section{Introduction}
\label{Sect:1}

\IEEEPARstart{D}{eep} neural networks (DNNs) have achieved remarkable success in various vision tasks, including autonomous driving, intelligent surveillance, and remote sensing image understanding. Despite these advancements, such models exhibit an inherent sensitivity to adversarial examples \cite{szegedy2014intriguing,goodfellow2015explaining}. Carefully crafted and often imperceptible perturbations can induce incorrect predictions. This vulnerability raises security concerns in real-world scenarios. Specifically, remote sensing object detection \cite{wang2025smallobject,wang2025aerialdetect} is increasingly utilized in critical applications such as land monitoring and national defense. Consequently, research on adversarial attacks against remote sensing object detectors is particularly important.

Adversarial attacks can be categorized into digital and physical domains based on the space in which perturbations are applied. Digital attacks directly manipulate input images. Physical attacks involve placing perturbations within real-world scenes followed by image re-acquisition for evaluation. Compared with digital attacks, physical attacks are more consistent with real-world deployment. Nevertheless, ensuring both attack effectiveness and physical feasibility remains a significant challenge.

Physical adversarial attacks are further divided into contact-based and contactless categories. Contact-based attacks apply perturbations directly to the target surface. These perturbations are relatively easy to deploy but visually conspicuous. Within this category, patch-based attacks \cite{thys2019fooling,brown2018adversarial,wang20253Dpatch} mislead DNNs by attaching adversarial patches, while camouflage-based attacks \cite{wang2021dual,wang2022fca,liu2026naturalistic} alter the overall appearance of the target. Compared with patch-based attacks, camouflage-based methods often achieve more stable performance under multiple viewpoints and complex observation conditions.
Contactless attacks interfere with the model perception process through optical devices, offering superior concealment. However, they are more sensitive to environmental conditions and often less stable in practice. Therefore, achieving strong attack performance while maintaining physical feasibility and concealment remains a key challenge in physical adversarial attack research.

Adversarial attacks are generally classified into white-box and black-box categories based on model accessibility. In white-box attacks, full access to the model architecture and parameters is assumed. It allows adversarial perturbations to be optimized directly through backpropagation. Nevertheless, such an assumption is often difficult to satisfy in practical applications. By contrast, black-box attacks are conducted with access only to model outputs through queries \cite{papernot2017practical}. Under this constraint, perturbations have to be optimized by estimating the model response from repeated interactions. While the unavailability of internal model information complicates the optimization process, black-box attacks are more consistent with real deployment conditions and possess greater practical importance.

Although black-box adversarial attacks have been extensively studied, existing research remains primarily concentrated on image classification. By comparison, black-box attacks against object detectors have received significantly less attention. Furthermore, existing physical attack methods are evaluated predominantly in the digital domain, leaving validation in real physical environments insufficient. This limitation is more pronounced in remote sensing scenarios due to substantial scale variations and complex imaging conditions. Consequently, black-box physical adversarial attacks against remote sensing object detectors deserve further investigation.

Motivated by prior studies on black-box attacks for image classification \cite{su2019onepixel,wang2023rfla}, this paper proposes ColorFD, a black-box physical patch attack against remote sensing object detectors. Compared with pixel-wise patch-texture optimization, ColorFD models adversarial perturbations as multiple small pure-color square patches. This representation reduces the number of decision variables and facilitates printing and physical deployment. The positions and color parameters of these patches are jointly optimized using Differential Evolution (DE). The most closely related DE-based black-box patch attack, Bbox-Att \cite{tang2022bboxatt}, mainly focuses on patch-texture dimensionality reduction and detector-level fitness design. However, a single overall fitness value provides limited optimization guidance when multiple targets are attacked simultaneously. To address this limitation, ColorFD introduces a target-wise fitness and selection mechanism. This mechanism preserves target-specific advantageous variables through partial component exchange. Regarding access to the target model, ColorFD requires only the returned detection results. It does not require model gradients or intermediate features. Therefore, ColorFD provides a practical robustness-evaluation approach for optical remote sensing detection systems under restricted model access.

The main contributions of this paper are summarized as follows.

\begin{itemize}
	\item [(1)]
	A physically realizable adversarial design based on multiple small pure-color square patches is proposed. These patches are easy to fabricate and deploy in real-world environments while maintaining strong attack effectiveness.
	
	\item [(2)]
	A target-wise fitness and selection mechanism is developed for multi-target black-box attacks. The target-level fitness evaluates the attack state of each target separately. The selection mechanism preserves target-level improvements and alleviates optimization conflicts across targets.
	
	\item [(3)]
	Two guidance strategies are introduced to constrain the patch search space. The key-region strategy uses finite-difference color probing to identify instance-specific sensitive regions. The common-feature strategy constructs a category-level prior that reduces repeated key-region localization while maintaining competitive attack performance.
	
\end{itemize}

\section{Related Work}
\label{Sect:2}

\textbf{Black-Box Attack.}
In a black-box attack, the attacker cannot access the internal architecture, parameters, or gradients of the target model. The attacker can only obtain outputs by querying the model. Under this setting, attack generation is typically formulated as a query-based optimization problem. Existing methods are broadly categorized into transfer-based attacks \cite{gan2025transferability} and query-based attacks \cite{bai2023query}.

Transfer-based attacks typically involve training a surrogate model to approximate the target model. Adversarial examples are then generated via white-box attacks on the surrogate and transferred to the target. While this design avoids frequent queries, its performance is strongly contingent upon the similarity between the surrogate and the target models.
In contrast, query-based attacks interact directly with the target model to optimize perturbations based on model outputs. Certain methods rely on finite-difference strategies \cite{chen2017zoo}, while others adopt evolutionary algorithms such as Genetic Algorithms \cite{alzantot2019genattack} and Particle Swarm Optimization (PSO) \cite{ma2026pso}. Beyond evolutionary algorithms, reinforcement learning \cite{wei2023jointlyopt} has also been explored for query-based black-box patch attacks. Although these methods are flexible and broadly applicable, they typically require a large number of model queries. Such a requirement limits their efficiency in practical applications.

\textbf{Patch Attack.}
Pixel-level perturbations utilized in digital attacks are difficult to preserve under physical transformations. Adversarial patches are regarded as a more practical alternative. These patches usually take the form of localized patterns attached to target objects to mislead the model. To date, patch-based attacks have been extensively explored across various vision tasks.

In face recognition, it was demonstrated by Sharif et al. \cite{sharif2016accessorize,sharif2019framework} that adversarial eyeglasses could deceive recognition systems. Similarly, facial stickers were designed by Wei et al. \cite{wei2022sticker} to attack FaceNet \cite{schroff2015facenet}. In pedestrian detection, Thys et al. \cite{thys2019fooling} extended AdvPatch \cite{brown2018adversarial} to attack YOLOv2 \cite{redmon2017yolo9000}. Furthermore, naturalistic T-shirt patterns were proposed by Xu et al. \cite{xu2020tshirt}, with their efficacy validated in physical environments. In traffic sign recognition, Chen et al. \cite{chen2018shapeshifter} introduced Shapeshifter to attack Faster R-CNN \cite{ren2015faster}. By sampling diverse physical conditions, this method generated robust adversarial patches under varying viewpoints and distances.
Despite their effectiveness, adversarial patches are often visually conspicuous. This high visibility limits their practical applicability. To alleviate this issue, Generative Adversarial Networks (GANs) \cite{goodfellow2014gan} were introduced to synthesize more natural-looking patches \cite{doan2022tnt}. Additionally, visual realism was improved through techniques such as knowledge distillation \cite{liu2025distillation} and style transfer \cite{duan2020camouflage}. Such efforts significantly enhanced both the stealthiness and usability of adversarial patches.

\textbf{Adversarial Attacks in Remote Sensing.}
Adversarial attacks on remote sensing images have attracted increasing attention in recent years. Within the digital domain, Wasserstein GAN \cite{arjovsky2017wgan} was utilized by Burnel et al. \cite{burnel2021remote} to generate visually natural adversarial examples. This method achieved promising results on multiple classification models. Similarly, a black-box attack was proposed by Xu et al. \cite{xu2022universal}. It utilized shallow features of surrogate models to identify common vulnerabilities, thereby enabling the discovery of cross-network weaknesses. Its effectiveness has been further validated in both scene classification and semantic segmentation tasks.
In the physical domain, a deployable multi-patch adversarial attack was introduced by Huang et al. \cite{huang2024dempaa} for remote sensing image classification. Regarding object detection, adversarial patches for multi-scale objects were developed by Zhang et al. \cite{zhang2022uav}. Furthermore, an adaptive patch attack for aircraft detection was proposed by Lian et al. \cite{lian2022benchmarking}. This method demonstrated strong robustness when patches were deployed both inside and outside target objects. Sun et al. \cite{sun2023tpa} proposed TPA, which employs first-order difference masking to select attacked sub-patches and uses a bounding-box drifting objective.
Despite these advancements, balancing attack effectiveness with physical feasibility remains a primary challenge in the field of remote sensing adversarial research.

\section{Preliminaries}
\label{Sect:3}
\subsection{Problem Definition}

Given an input image $x$, an object detector outputs a set of detections:
\begin{equation}
	\mathcal{D}(x)=\left\{d_t\right\}_{t=1}^{T}=\left\{(b_t,p_t)\right\}_{t=1}^{T},
\end{equation}
where $d_t$ represents the detection indexed by $t$. Specifically, $b_t$ denotes the corresponding bounding box, $p_t$ denotes the associated class probability distribution, and $T$ signifies the total number of detections. The primary objective of an adversarial attack involves transforming $x$ into an adversarial example $x^{adv}$ to induce incorrect detection results.

In this paper, a patch-based adversarial attack is considered. Multiple patches are applied to each selected target to achieve object-level disappearance. The adversarial patch set is defined as
\begin{equation}
	\mathcal{P}=\{P_t\}_{t=1}^{T}, \quad P_t=\{P_{t,n}\}_{n=1}^{N_t},
\end{equation}
where $P_t$ represents the patch set associated with the target indexed by $t$, and $P_{t,n}$ denotes the patch indexed by $n$ within this set. The variable $N_t$ signifies the total number of patches applied to that target. Each patch $P_{t,n}$ is parameterized by a patch parameter vector $z_{t,n}$ and a size parameter $s_t$ specific to the target indexed by $t$. The adversarial example is constructed as follows:
\begin{equation}
	x^{adv} = x \odot (1 - M) + \sum_{t=1}^{T} \sum_{n=1}^{N_t} M_{t,n} \odot P_{t,n},
\end{equation}
where $M_{t,n}$ denotes the binary mask corresponding to patch $P_{t,n}$, and $M$ denotes the binary mask covering all applied patches. In this formulation, $\odot$ represents the element-wise multiplication operator.

Object disappearance is adopted as the attack success criterion. For the target indexed by $t$, the attack is considered successful if no valid prediction of the same class is produced. It is also successful when all predicted bounding boxes of that class have the Intersection over Union (IoU) with ground-truth box below a predefined threshold. Formally, the success indicator is defined as:
\begin{equation}
	S_t =
	\begin{cases}
		1, & \mathcal{B}_t = \varnothing \ \text{or} \ \max\limits_{b_j \in \mathcal{B}_t} \mathrm{IoU}(b_j, b_t^{gt}) < \tau, \\
		0, & \text{otherwise},
	\end{cases}
\end{equation}
where $\mathcal{B}_t$ represents the set of predicted bounding boxes with the same class as the target indexed by $t$, $b_t^{gt}$ denotes the ground-truth bounding box, and $\tau$ signifies the IoU threshold. Under this definition, the attack is regarded as successful when $S_t$ = 1.

The patch set $\mathcal{P}$ is optimized to suppress valid detections of the selected targets. This objective is formulated as the following optimization problem:
\begin{equation}
	\mathcal{P}^{*} = \arg\min_{\mathcal{P}} \mathcal{F}(x^{adv}),
\end{equation}
where $\mathcal{F}(\cdot)$ represents the fitness function evaluated on the detector outputs. This formulation allows the optimal patch parameters to be determined by minimizing the response of the target detector.

\subsection{Differential Evolution}
DE is a population-based optimization algorithm designed to iteratively update a set of candidate solutions toward the global optimum. Since DE operates without the requirement for gradient information, it is well suited for black-box optimization scenarios.

Let the population be denoted by
\begin{equation}
	\mathcal{X}^g = \left\lbrace{x_i^g}\right\rbrace_{i=1}^{N_p},
\end{equation}
where $x_i^g$ represents the individual indexed by $i$, and $g$ signifies the generation index. The variable $N_p$ denotes the total population size. Within this framework, each individual $x_i^g$ corresponds to a candidate adversarial patch set $\mathcal{P}_i^g$.

In the mutation step, the rand-to-best/1 strategy is adopted:
\begin{equation}
	v_i^g = x_{r1}^g + F \cdot (x_{best}^g - x_i^g) + F \cdot (x_{r2}^g - x_{r3}^g),
\end{equation}
where $v_i^g$ represents the mutant vector associated with the individual indexed by $i$, and $F$ signifies the scaling factor. In this context, $x_{best}^g$ denotes the optimal individual within the current population. $x_{r1}^g$ and $x_{r2}^g$ represent two randomly selected individuals used to incorporate stochastic perturbations.
After mutation, crossover is performed to generate a trial vector:
\begin{equation}
	u_{i,j}^g =
	\begin{cases}
		v_{i,j}^g, & \text{if } r_j \leq CR \ \text{or } j = j_{rand}, \\
		x_{i,j}^g, & \text{otherwise},
	\end{cases}
\end{equation}
where $u_{i,j}^g$ represents the component of the trial vector $u_i^g$ indexed by $j$. The parameter $CR$ signifies the crossover rate, and $r_j$ is sampled from a uniform distribution over $(0,1)$. In addition, the index $j_{rand}$ ensures that at least one dimension is inherited from the mutated vector. Following the crossover, the trial vector is evaluated. Between $u_i^g$ and $x_i^g$, the individual exhibiting superior fitness is selected for the next generation.

When applying DE to adversarial patch generation for object detection, two primary challenges arise. First, the attack on multiple targets complicates individual evaluation. A single evaluation metric often fails to provide optimization information across diverse target states. Second, defining the search space over entire bounding boxes leads to high dimensionality. This excessive search space significantly reduces optimization efficiency. To address multi-target evaluation, a tailored fitness function and a target-wise selection mechanism are designed. Optimization efficiency is further improved by introducing key-region and common-feature guidance to constrain the search space. Detailed descriptions of these components are provided in Section~\ref{Sect:4}.

\section{Method}
\label{Sect:4}
\begin{figure*}[!htb]
	\centering
	\includegraphics[width=1.5\columnwidth]{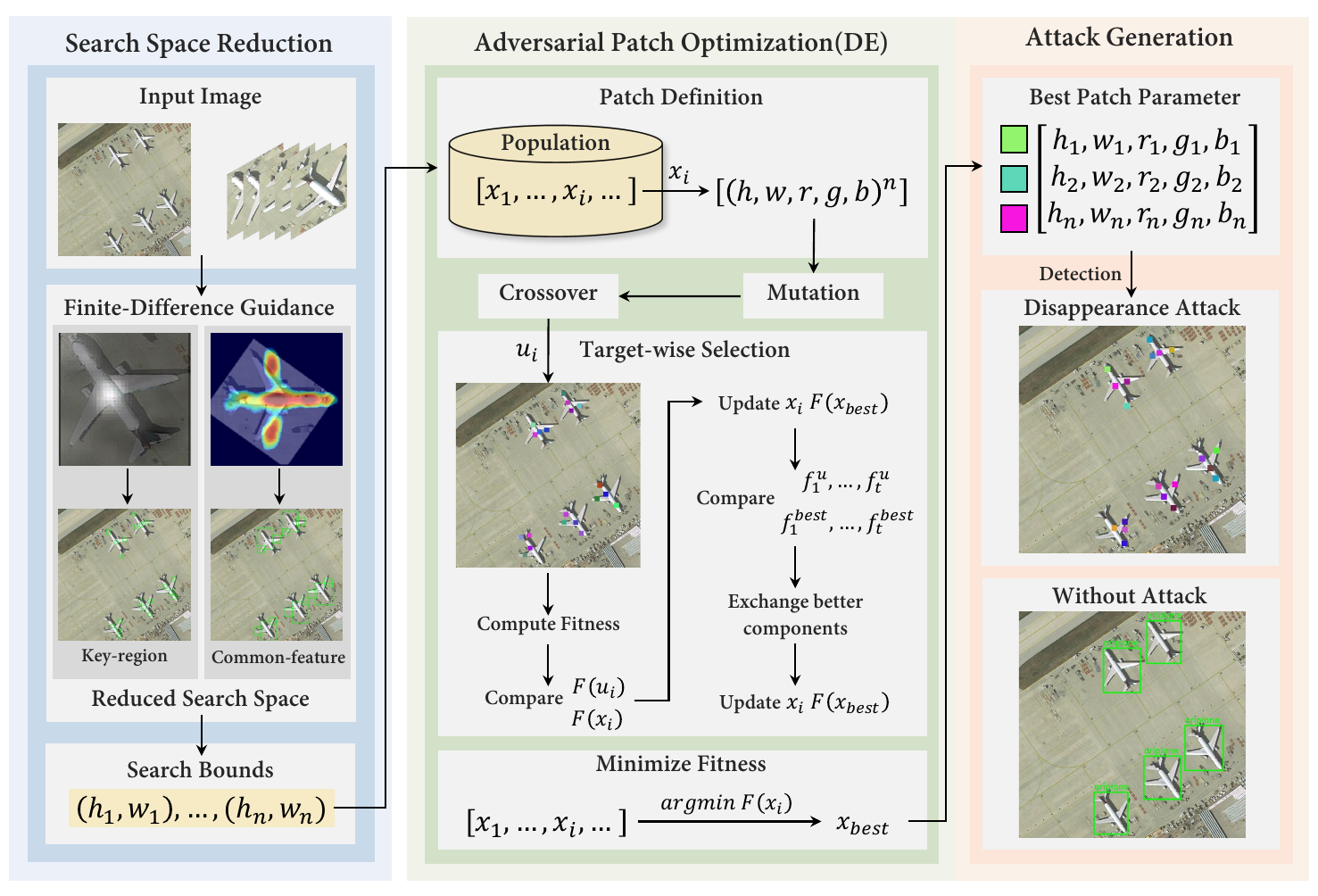}
	\caption{Overview of the proposed ColorFD framework. The method first localizes key regions through finite-difference analysis to constrain the search space, and then optimizes patch parameters using differential evolution with the proposed fitness and selection strategies.}
	\label{fig:1}
\end{figure*}

This section details the proposed ColorFD attack through two dimensions: the evolutionary mechanism and search space reduction. The overall pipeline is illustrated in Fig.~\ref{fig:1}.

\subsection{Fitness Function}
\label{4.1}

Within the proposed attack framework, each individual encodes patch parameters for multiple targets simultaneously. Optimization guided solely by the overall fitness of an individual fails to provide optimization state for all targets. This limitation reduces search efficiency and increases the risk of converging to local optima. In response to this deficiency, a fitness function incorporating target-level evaluation is introduced into the DE process. This design allows both the attack state of each target and the individual as a whole to be evaluated. Consequently, a more informative optimization signal is provided for multi-target attacks.

For the target indexed by $t$, the target-level fitness of individual $x_i^g$ is defined as:
\begin{equation}
	f_t(x_i^g) = \max_{d_j \in \mathcal{D}(x_i^g)} S(d_j),
\end{equation}
where $\mathcal{D}(x_i^g)$ represents the set of detections obtained from the adversarial example generated by $x_i^g$. The scoring function $S(\cdot)$ is formulated as:
\begin{equation}
	S(d_j) =
	\begin{cases}
		s_j, & \text{if } c_j = c_t^{gt} \ \text{and } \mathrm{IoU}(b_j, b_t^{gt}) \geq \tau, \\
		0, & \text{otherwise},
	\end{cases}
\end{equation}
where $c_j$ and $s_j$ signify the predicted class label and confidence score of detection $d_j$, respectively. The variable $c_t^{gt}$ denotes the ground-truth class label of the target indexed by $t$. The overall fitness of individual $x_i^g$ is then defined as:
\begin{equation}
	\mathcal{F}(x_i^g) = \sum_{t=1}^{T} f_t(x_i^g).
\end{equation}

The optimization process minimizes the fitness value, where a lower value signifies a weaker detector response to the target objects. This reduction directly corresponds to a stronger disappearance effect. By suppressing the maximum valid detection confidence for each target, this fitness design encourages the simultaneous disappearance of all selected targets.

\subsection{Target-wise Selection Mechanism}
\label{4.2}
To exploit target-level fitness, a target-wise selection mechanism is introduced. During evolution, each offspring is first compared with its parent based on their overall fitness. If the offspring achieves a lower overall fitness, it directly replaces the parent. Otherwise, a target-wise comparison is performed to identify specific targets where the offspring yields no higher target-level fitness. For these targets, the corresponding decision variables in the parent are partially replaced to form an exchanged individual. This new individual is then evaluated, and the update is accepted only if its fitness is no higher than that of the original parent.

This process is applied sequentially to all targets. It preserves target-level improvements while maintaining the overall fitness of the individual. By jointly considering global and target-level fitness information, the proposed mechanism alleviates optimization conflicts across targets and supports more efficient multi-target optimization. The detailed procedure is outlined in Algorithm~\ref{alg:target_wise_selection}.

\begin{algorithm}[!htb]
	\caption{Target-wise Selection Mechanism}
	\label{alg:target_wise_selection}
	\begin{algorithmic}[1]
		\REQUIRE Parent individual $x_i^g$, offspring $u_i^g$, target-level fitness $\{f_t(\cdot)\}_{t=1}^{T}$, overall fitness $\mathcal{F}(\cdot)$
		\ENSURE Updated individual $x_i^{g+1}$
		\IF{$\mathcal{F}(u_i^g) < \mathcal{F}(x_i^g)$}
		\STATE $x_i^{g+1} \leftarrow u_i^g$
		\ELSE
		\STATE $x_i^{g+1} \leftarrow x_i^g$
		\FOR{$t = 1$ to $T$}
		\IF{$f_t(u_i^g) \leq f_t(x_i^{g+1})$}
		\STATE Construct $\hat{x}_i^g$ by replacing the variables of target $t$ in $x_i^{g+1}$ with those from $u_i^g$
		\IF{$\mathcal{F}(\hat{x}_i^g) \leq \mathcal{F}(x_i^{g+1})$}
		\STATE $x_i^{g+1} \leftarrow \hat{x}_i^g$
		\ENDIF
		\ENDIF
		\ENDFOR
		\ENDIF
		\RETURN $x_i^{g+1}$
	\end{algorithmic}
\end{algorithm}

\subsection{Key-Region Localization}
\label{4.3}
As the number of patch variables increases, the search space grows rapidly, making DE optimization computationally expensive. Since each patch is jointly determined by its position and color, direct optimization over the entire bounding box results in high dimensionality.

\begin{figure*}[!htb]
	\centering
	\includegraphics[width=1.8\columnwidth]{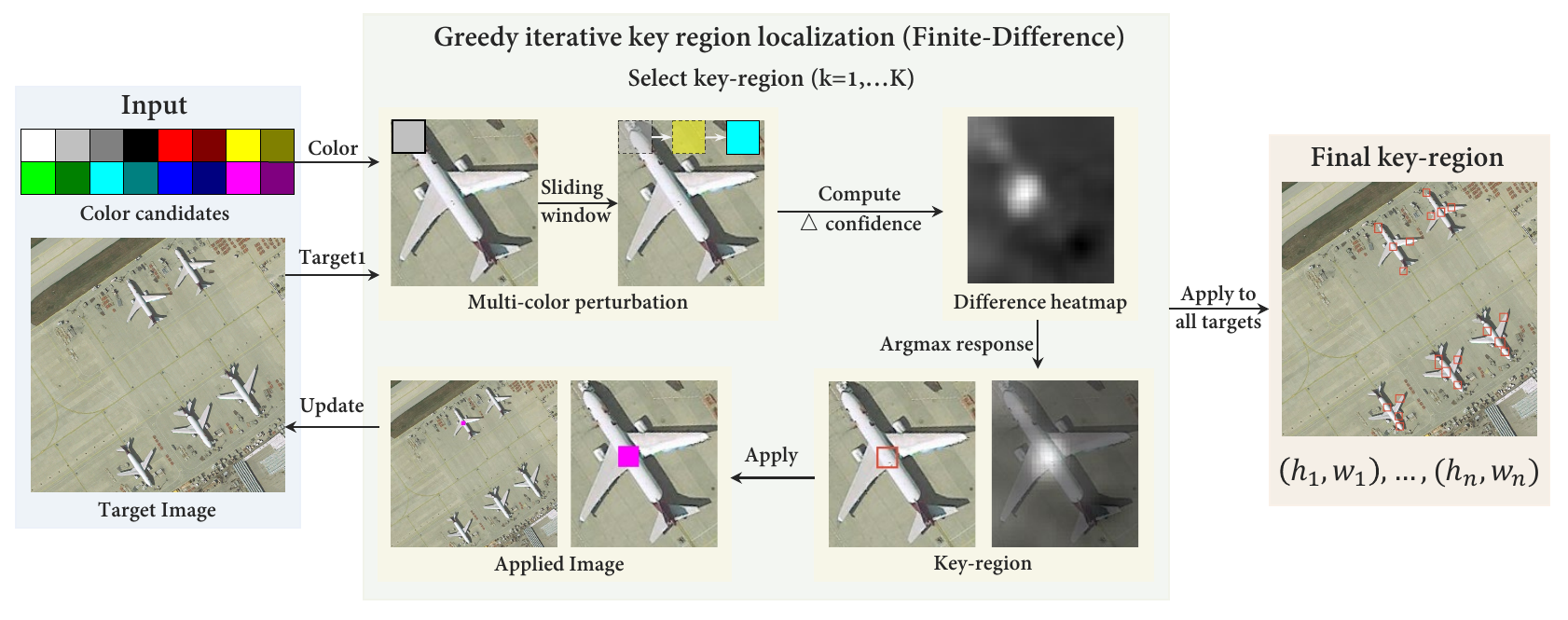}
	\caption{Pipeline of the proposed key-region localization process based on greedy finite-difference guidance. Key regions are iteratively identified by evaluating confidence variations via multi-color perturbations and sliding windows. This localization procedure is applied to all instances within the image sequentially.}
	\label{fig:2}
\end{figure*}

To mitigate this, a finite-difference-based key-region localization method is proposed to identify sensitive regions and constrain the search space. Specifically, the ground-truth bounding box serves as the initial search area. Within this region, a sliding window traverses candidate locations to apply pure-color perturbations from a predefined set. The resulting detector response is recorded as a finite-difference signal to generate a response map. Based on this map, the method identifies areas with strong responses and selects the most prominent connected component as a key region. Once the corresponding region-color pair is applied, the search continues on the updated image. This greedy process sequentially identifies multiple key regions, thereby progressively restricting the search space for subsequent optimization. Such a constrained approach ensures that the evolutionary search is concentrated on high-impact areas rather than the entire bounding box. The overall procedure is illustrated in Fig.~\ref{fig:2} and Algorithm~\ref{alg:key-region}.

\begin{algorithm}[!htb]
	\caption{Key-Region Localization via Finite-Difference Color Probing}
	\label{alg:key-region}
	\begin{algorithmic}[1]
		\REQUIRE Input image $x$, target box $b_t$, detector $D(\cdot)$, color set $\mathcal{C}$, number of key regions $K$
		\ENSURE Key region set $\mathcal{R}$
		\STATE $\mathcal{R} \leftarrow \emptyset$, \quad $\tilde{x} \leftarrow x$
		\FOR{$k = 1$ to $K$}
		\STATE Initialize response map $M$
		\FOR{each sliding-window position $w \in b_t$}
		\FOR{each color $c \in \mathcal{C}$}
		\STATE Generate perturbed image $\hat{x}$ by applying color $c$ at position $w$ on $\tilde{x}$
		\STATE Compute finite-difference response $\Delta(w,c)$ from the confidence change before and after perturbation
		\STATE Update $M$ with $\Delta(w,c)$
		\ENDFOR
		\ENDFOR
		\STATE Select the next key region $r_k$ and color $c_k$ according to $M$
		\STATE $\mathcal{R} \leftarrow \mathcal{R} \cup \{r_k\}$
		\STATE Update $\tilde{x}$ by applying $(r_k, c_k)$
		\ENDFOR
		\RETURN $\mathcal{R}$
	\end{algorithmic}
\end{algorithm}

\subsection{Common-Feature Extraction}
\label{4.4}
Although key-region localization improves search efficiency, the exhaustive traversal of candidate regions remains computationally expensive. Targets belonging to the same category often exhibit consistent spatial patterns within their key regions. Motivated by this observation, a common-feature extraction method is proposed to aggregate key-region distributions across instances of the same category. This approach allows redundant computations to be minimized by leveraging category-level prior information. Specifically, target regions are first cropped and normalized into a unified coordinate system. Pose-based alignment, scale normalization, and padding are then applied to project targets from diverse samples into a common reference space. Following this alignment, the localized key regions are converted into binary masks and smoothed through Gaussian filtering to produce continuous response distributions. For each individual target, the masks of all localized regions are aggregated to form a target-level heatmap. The method then accumulates and normalizes these heatmaps across all instances of the same category to generate a class-level common-feature heatmap. Based on this heatmap, the $K$ most prominent connected components characterized by strong responses are extracted as common-feature regions. These regions directly guide the search space of DE-based patch optimization. Unlike key-region localization, the proposed method operates at the category level. This design significantly reduces computational cost while providing robust prior guidance for the subsequent optimization process. The procedure of this mechanism is presented in Algorithm~\ref{alg:common-feature}.

\begin{algorithm}[!htb]
	\caption{Common-Feature Extraction}
	\label{alg:common-feature}
	\begin{algorithmic}[1]
		\REQUIRE Target set $\mathcal{T}$ of the same category, target boxes $\mathcal{B}$, key regions $\{\mathcal{R}_t\}$, pose angles $\Theta$, number of common regions $K$
		\ENSURE Common-Feature region set $\mathcal{R}^{c}$
		\STATE Initialize common-feature heatmap $\mathcal{H} \leftarrow 0$
		\FOR{each target $t \in \mathcal{T}$}
		\STATE Crop target region according to $b_t \in \mathcal{B}$
		\STATE Normalize coordinates of key regions $\mathcal{R}_t$
		\STATE Align target and key regions using $\theta_t \in \Theta$
		\STATE Resize and pad the aligned target to a unified canvas
		\STATE Initialize target heatmap $\mathcal{H}_t \leftarrow 0$
		\FOR{each key region $r \in \mathcal{R}_t$}
		\STATE Convert $r$ into a binary mask $M_r$
		\STATE Apply Gaussian smoothing to $M_r$
		\STATE $\mathcal{H}_t \leftarrow \mathcal{H}_t + M_r$
		\ENDFOR
		\STATE $\mathcal{H} \leftarrow \mathcal{H} + \mathrm{MinMaxNorm}(\mathcal{H}_t)$
		\ENDFOR
		\STATE $\mathcal{H} \leftarrow \mathrm{MinMaxNorm}(\mathcal{H})$
		\STATE Extract the top-$K$ most prominent connected components from $\mathcal{H}$ as $\mathcal{R}^{c}$
		\RETURN $\mathcal{R}^{c}$
	\end{algorithmic}
\end{algorithm}

\section{Experiments}
\label{Sect:5}
\subsection{Experimental Settings}

\textbf{Dataset.}
%Experiments are conducted on the DIOR \cite{li2020dior} remote sensing object detection dataset. Our method is not limited to a single target. Because aircraft features are well-visible, a subset of aircraft is constructed from the DIOR to demonstrate the experimental results. The subset contains 100 images and 407 airplane instances. It covers multiple aircraft types, including jet aircraft, turboprop aircraft, and piston-engine aircraft. Object scale varies from approximately $30 \times 30$ to $300 \times 300$ pixels. Diverse pose distributions are also included. This diversity provides a suitable basis for evaluating the proposed method under complex remote sensing conditions.
Experiments are conducted on the DIOR remote sensing object detection dataset \cite{li2020dior}. The proposed method is applicable to multiple target categories rather than being restricted to aircraft. Aircraft are selected for evaluation because they are representative targets in remote sensing imagery and commonly exhibit substantial variations in scale, orientation and appearance. Their distinctive structures also allow the attack effects to be assessed clearly and reliably in both quantitative and qualitative analyses. Accordingly, an aircraft subset is constructed from DIOR for the experiments. The subset contains 100 images and 407 aircraft instances. It includes several aircraft types, such as jet, turboprop, and piston-engine aircraft. The object sizes range from approximately $30 \times 30$ to $300 \times 300$ pixels. The aircraft instances also exhibit diverse orientations and poses. These variations provide a suitable basis for evaluating the proposed method under complex remote sensing conditions.

\textbf{Evaluation Metrics.}
Object disappearance is adopted as the attack objective. Attack Success Rate (ASR) is used as the primary metric for target-level attack evaluation. AP@0.5 (AP$_{50}$) is further introduced to assess the overall degradation of detection performance. Together, ASR and AP$_{50}$ provide a more comprehensive evaluation of attack effectiveness.

\textbf{Implementation Details.}
The proposed method is evaluated on YOLOv3u, YOLOv5u, and Faster R-CNN. YOLOv3u and YOLOv5u are implemented based on the Ultralytics framework, while Faster R-CNN is implemented using the MMDetection framework. All models are initialized with pretrained weights and fine-tuned on the DIOR dataset for 50, 50, and 36 epochs, respectively. The best checkpoints are selected for subsequent evaluation.
During inference, all models use a unified input resolution of $800 \times 800$ and a confidence threshold of $0.3$. The same settings are used for both detection and evaluation. For the DE algorithm, the population is initialized randomly. The main hyperparameters are set as follows: the population size multiplier is 3, the maximum number of iterations is 100, and the patch ratio is 0.15. Each target is assigned four pure-color patches for optimization. The IoU matching threshold and the key-region selection threshold are set to 0.3 and 0.85, respectively. All experiments are conducted on a single NVIDIA GeForce RTX 4090 GPU with 48 GB memory.

\textbf{Comparison Methods and Implementation Details.}
The proposed method is compared against the black-box method Bbox-Att and white-box baselines AP-PA and BADEI \cite{wang2023badei}. All methods are implemented using their official codebases.
To ensure a fair comparison, key perturbation parameters are standardized across all methods to maintain comparable perturbation magnitudes. Specifically, in Bbox-Att, the sub-patch size and patch dimension are set to 10 and 40, respectively. In AP-PA and BADEI, the patch ratios are set to 0.5 and 0.11, respectively. All methods are evaluated under identical settings.

\subsection{Digital Adversarial Attack}

\subsubsection{Effectiveness of Key-Region Guidance}
\begin{figure}[!htb]
	\centering
	\subfloat[]{%
		\includegraphics[width=0.9\columnwidth]{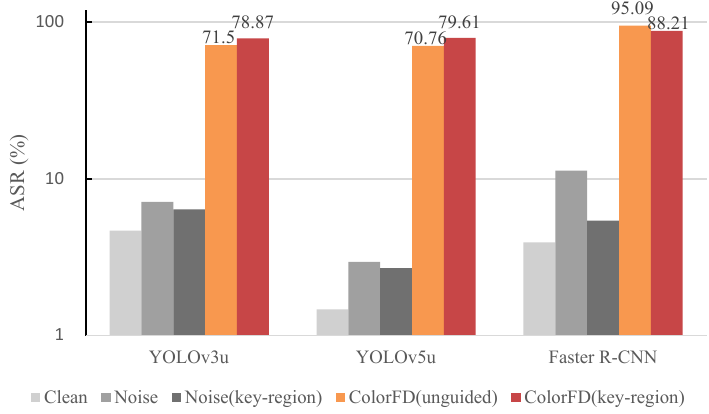}
		\label{fig:3.a}}\\[1mm]
	\subfloat[]{%
		\includegraphics[width=0.9\columnwidth]{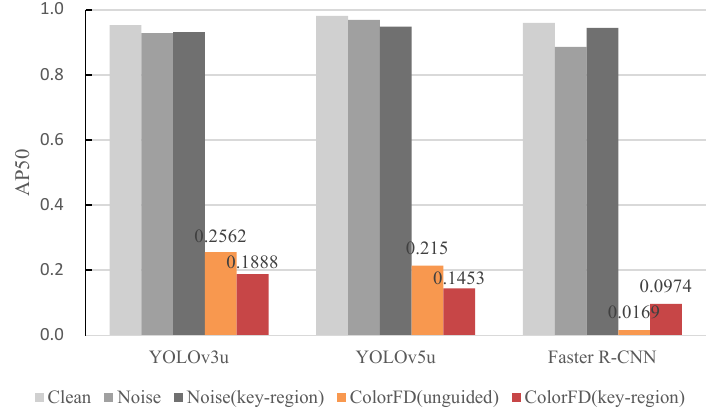}
		\label{fig:3.b}}
	\caption{Performance comparison of unguided and key-region-guided ColorFD against noise-based baselines across three object detectors.
		(a) ASR comparison.
		(b) AP$_{50}$ comparison.}
	\label{fig:3}
\end{figure}

The key-region guidance strategy is employed to mitigate the challenges associated with DE in high-dimensional search spaces. The corresponding effectiveness is examined through evaluations in the digital domain. Fig.~\ref{fig:3} and Table~\ref{tab1:digital_attack_results} provide quantitative comparisons across various detectors.

\begin{table*}[t]
	\centering
	\caption{Comparison of digital-space attack results obtained
		using different methods on three object detectors. Higher ASR and lower AP$_{50}$ indicate stronger attack performance.
		Bold values indicate the best result for each detector and metric.}
	\label{tab1:digital_attack_results}
	
	\setlength{\tabcolsep}{5pt}
	\begin{tabular*}{\textwidth}
		{@{\extracolsep{\fill}}lcccccc}
		\toprule
		\multirow{2}{*}{Method}
		& \multicolumn{2}{c}{YOLOv3u}
		& \multicolumn{2}{c}{YOLOv5u}
		& \multicolumn{2}{c}{Faster R-CNN} \\
		\cmidrule(lr){2-3}
		\cmidrule(lr){4-5}
		\cmidrule(lr){6-7}
		& ASR (\%) & AP$_{50}$
		& ASR (\%) & AP$_{50}$
		& ASR (\%) & AP$_{50}$ \\
		\midrule
		Clean
		& 4.67 & 0.9529
		& 1.47 & 0.9816
		& 3.93 & 0.9599 \\
		
		Noise
		& 7.13 & 0.9282
		& 2.95 & 0.9688
		& 11.30 & 0.8860 \\
		
		Noise (key-region)
		& 6.39 & 0.9314
		& 2.70 & 0.9482
		& 5.41 & 0.9448 \\
		
		ColorFD (unguided)
		& 71.50 & 0.2562
		& 70.76 & 0.2150
		& \textbf{95.09} & \textbf{0.0169} \\
		
		ColorFD (key-region)
		& \textbf{78.87} & \textbf{0.1888}
		& \textbf{79.61} & \textbf{0.1453}
		& 88.21 & 0.0974 \\
		\bottomrule
	\end{tabular*}
	
%	\vspace{2pt}
%	\parbox{\textwidth}{\footnotesize
%		\textit{Note:} ASR denotes attack success rate.
%		AP$_{50}$ denotes average precision at an IoU threshold of 0.5.
%		Higher ASR and lower AP$_{50}$ indicate stronger attack performance.
%		Bold values indicate the best result for each detector and metric.
%	}
\end{table*}

As illustrated in Fig.~\ref{fig:3} and Table~\ref{tab1:digital_attack_results}, all detectors maintain robust performance on clean images. These models exhibit low ASR and high AP$_{50}$ under normal conditions. While random and key-region noise cause only marginal performance degradation, ColorFD leads to a substantial increase in ASR. Concurrently, a significant reduction in AP$_{50}$ is observed across the models.

Regarding YOLOv3u and YOLOv5u, the unguided version of ColorFD achieves strong attack performance. The integration of key-region guidance further enhances both metrics. Such results indicate that the strategy successfully concentrates perturbations on regions critical for detection. On Faster R-CNN, however, the unguided ColorFD already reaches a high ASR. The introduction of key-region guidance does not yield further improvement and instead causes a minor performance drop. This phenomenon is likely attributed to the two-stage architecture of Faster R-CNN. In a black-box setting, this structure restricts access to intermediate information from the Region Proposal Network (RPN). Consequently, the finite-difference signals derived from final detection scores may provide less reliable guidance for localizing target-sensitive regions.

The quantitative results further substantiate these observations. For YOLOv3u and YOLOv5u, key-region guidance improves ASR by $7.37$ and $8.85$ percentage points, respectively. Meanwhile, AP$_{50}$ is reduced by $0.0674$ and $0.0697$. These findings demonstrate that key-region guidance achieves superior performance compared to the unguided variant on one-stage detectors under an identical patch budget.
In conclusion, key-region-guided ColorFD consistently outperforms both the unguided variant and noise-based baselines on one-stage detectors. Although its advantage is less pronounced on the two-stage detector, the overall effectiveness remains robust across all evaluated models.

\subsubsection{Effectiveness of Common-Feature Guidance}
\begin{figure}[!htb]
	\centering
	\includegraphics[width=0.85\columnwidth]{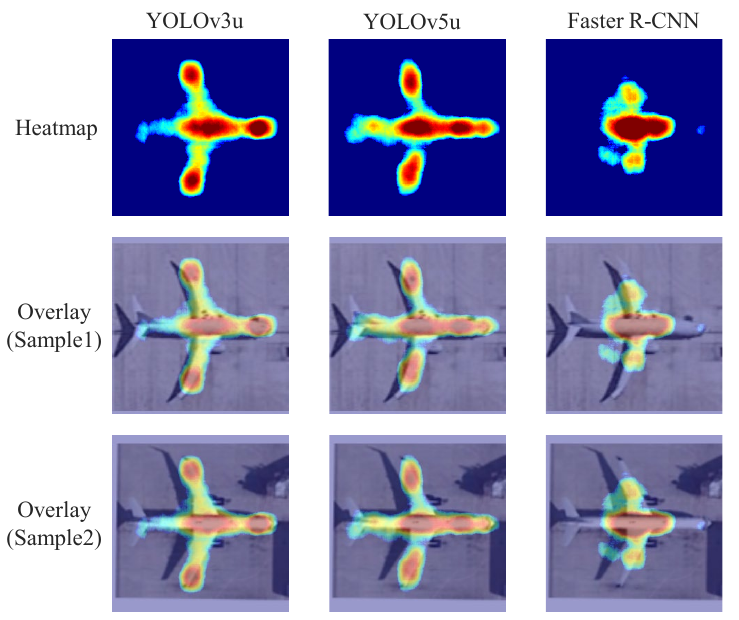}
	\caption{Common-Feature heatmaps and overlay visualizations across different object detectors. The heatmaps illustrate the frequency of key regions across various locations of the aircraft, primarily concentrated on the fuselage and the wings.}
	\label{fig:4}
\end{figure}

To evaluate the guiding capability of common-feature priors, experiments are conducted on a subcategory of dual-engine wing-mounted jet aircraft. A total of 306 instances are cropped, aligned, and normalized. Subsequently, a common-feature heatmap is constructed by aggregating the key-region distributions of these instances. The corresponding visualization results are presented in Fig.~\ref{fig:4}.

As illustrated in Fig.~\ref{fig:4}, the common-feature regions extracted by various detectors exhibit high spatial consistency. These regions are predominantly located around the wings and the front-central fuselage. This observation suggests that such areas not only play a critical role within individual models but also represent consistent cross-model semantic features. It further indicates that detectors generally depend on these structural components for object recognition.
Based on these findings, the extracted common-feature regions are incorporated as prior information to guide the ColorFD search process. Specifically, three patches are constrained to search within the identified common-feature regions. To maintain consistency with the unguided setting, the fourth patch is allowed to search freely within the bounding box.

\begin{table*}[t]
	\centering
	\caption{Comparison of digital-space attack results obtained
		using different guidance strategies on dual-engine wing-mounted
		jet aircraft across three object detectors. Higher ASR and lower AP$_{50}$ indicate stronger attack performance.
		Bold values indicate the best result for each detector and metric.}
	\label{tab2:digital_attack_common_results}
	
	\setlength{\tabcolsep}{5pt}
	\begin{tabular*}{\textwidth}
		{@{\extracolsep{\fill}}lcccccc}
		\toprule
		\multirow{2}{*}{Method}
		& \multicolumn{2}{c}{YOLOv3u}
		& \multicolumn{2}{c}{YOLOv5u}
		& \multicolumn{2}{c}{Faster R-CNN} \\
		\cmidrule(lr){2-3}
		\cmidrule(lr){4-5}
		\cmidrule(lr){6-7}
		& ASR (\%) & AP$_{50}$
		& ASR (\%) & AP$_{50}$
		& ASR (\%) & AP$_{50}$ \\
		\midrule
		Clean
		& 5.56 & 0.9438
		& 1.96 & 0.9755
		& 4.90 & 0.9498 \\
		
		Noise
		& 7.84 & 0.9208
		& 2.94 & 0.9683
		& 13.73 & 0.8613 \\
		
		Noise (key-region)
		& 6.86 & 0.9262
		& 2.61 & 0.9578
		& 6.54 & 0.9329 \\
		
		Noise (common-feature)
		& 6.86 & 0.9304
		& 2.61 & 0.9676
		& 19.28 & 0.7944 \\
		
		ColorFD (unguided)
		& 72.22 & 0.2411
		& 71.57 & 0.2037
		& 97.39 & 0.0045 \\
		
		ColorFD (common-feature)
		& 77.12 & \textbf{0.1912}
		& 75.16 & 0.1658
		& \textbf{99.35} & \textbf{0.0003} \\
		
		ColorFD (key-region)
		& \textbf{78.10} & 0.1949
		& \textbf{80.39} & \textbf{0.1459}
		& 87.91 & 0.0840 \\
		\bottomrule
	\end{tabular*}
	
%	\vspace{2pt}
%	\parbox{\textwidth}{\footnotesize
%		\textit{Note:} ASR denotes attack success rate.
%		AP$_{50}$ denotes average precision at an IoU threshold of 0.5.
%		Higher ASR and lower AP$_{50}$ indicate stronger attack performance.
%		Bold values indicate the best result for each detector and metric.
%	}
\end{table*}

The quantitative results are summarized in Table~\ref{tab2:digital_attack_common_results}. Compared with the unguided ColorFD, the common-feature guided variant achieves higher ASR and lower AP$_{50}$ across all three detectors. This performance demonstrates that the strategy provides effective spatial priors for enhancing attack effectiveness. On YOLOv3u and YOLOv5u, the performance of this variant is slightly inferior to key-region guidance but remains substantially better than the unguided baseline. Notably, on Faster R-CNN, common-feature guidance achieves superior performance, with ASR reaching $99.35\%$ and AP$_{50}$ approaching zero. This outcome effectively compensates for the limitations of key-region guidance when applied to two-stage detectors.

\begin{figure}[!htb]
	\centering
	\includegraphics[width=\columnwidth]{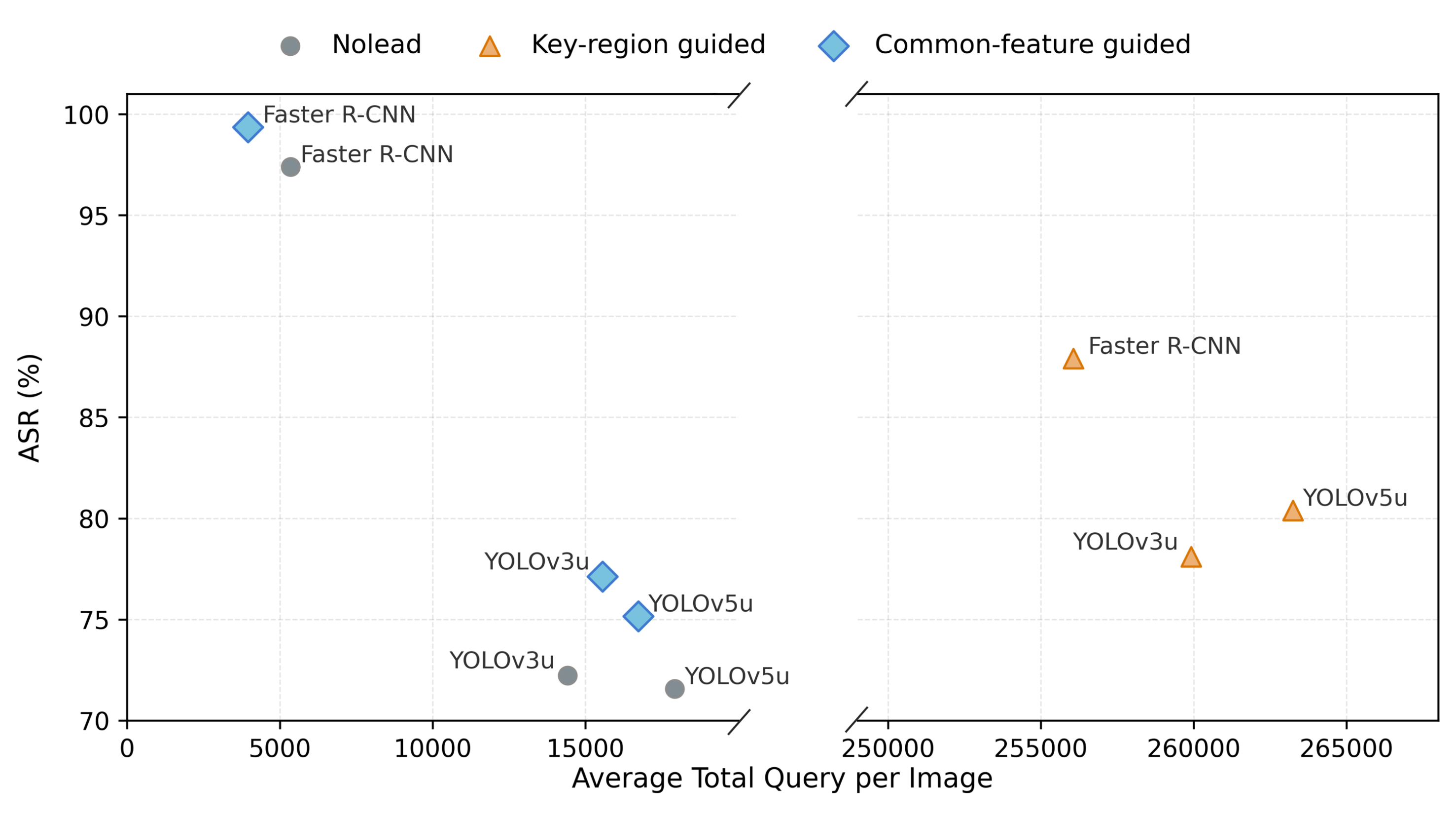}
	\caption{ASR versus average total query per image under different guidance strategies.
	Higher ASR and lower average total query per image under the same model indicate stronger attack performance.}
	\label{fig:5}
\end{figure}

Further analysis of query efficiency, as presented in Fig.~\ref{fig:5}, reveals distinct trade-offs between attack effectiveness and query cost under different guidance strategies. Key-region guidance exhibits strong attack performance. However, it requires a substantial number of queries. This is due to the necessity of performing finite-difference localization for each target individually. In contrast, common-feature guidance leverages precomputed class-level priors, thereby avoiding repeated per-target localization. Consequently, this approach attains competitive ASR with significantly fewer queries than key-region guidance while still outperforming the unguided method. These results substantiate that common-feature guidance offers a more favorable trade-off between attack effectiveness and query efficiency.

\begin{figure}[!htb]
	\centering
	\subfloat[]{%
		\includegraphics[width=\columnwidth]{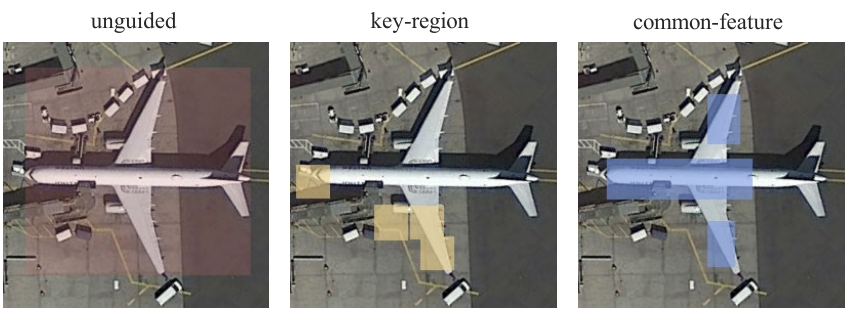}
		\label{fig:6.a}}\\[1mm]
	\subfloat[]{%
		\includegraphics[width=\columnwidth]{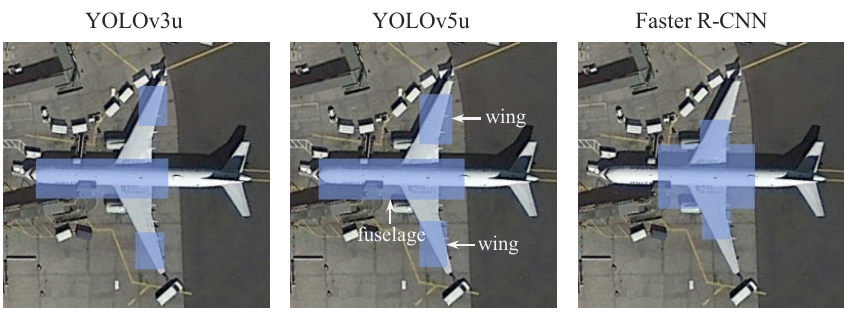}
		\label{fig:6.b}}
	\caption{Visualization of search space distributions under different guidance strategies.
		(a) Search regions under different guidance strategies. 
		(b) Consistent semantic regions under the common-feature strategy.}
	\label{fig:6}
\end{figure}

To further elucidate the efficiency discrepancy, Fig.~\ref{fig:6} visualizes the search space distributions under different guidance strategies. It is observed that the search space under common-feature guidance is more extensive than that under key-region localization. Nevertheless, this space remains substantially more constrained compared to the unguided configuration. Such a distribution demonstrates that common-feature guidance effectively restricts the search space. Moreover, these constrained regions align well with the semantic structure of the targets. This alignment suggests that the proposed method exploits cross-model semantic features to reduce the search space, thereby enhancing overall optimization efficiency.

In summary, although common-feature guidance is slightly less effective than key-region localization on one-stage detectors, it substantially reduces computational costs. The extracted semantic regions demonstrate that this method captures consistent discriminative features at the category level. This provides generalizable priors for black-box adversarial attacks.

\subsubsection{Comparison with Existing Methods}
\begin{figure}[!htb]
	\centering
	\subfloat[]{%
		\includegraphics[width=0.9\columnwidth]{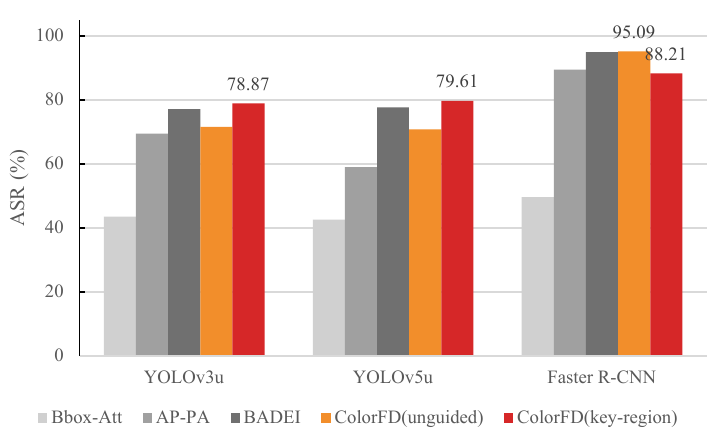}
		\label{fig:7.a}}\\[1mm]
	\subfloat[]{%
		\includegraphics[width=0.9\columnwidth]{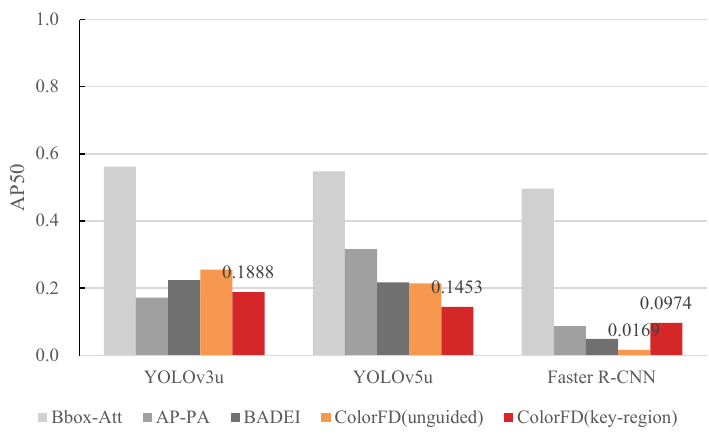}
		\label{fig:7.b}}
	\caption{Performance comparison of ColorFD and other attack methods on three object detectors.
		(a) ASR comparison.
		(b) AP$_{50}$ comparison.}
	\label{fig:7}
\end{figure}

To further validate the effectiveness of the proposed method, a comparative analysis is performed against Bbox-Att, AP-PA and BADEI. Under consistent perturbation budgets and experimental settings, the quantitative results are presented in Fig.~\ref{fig:7} and Table~\ref{tab3:comparison_attack_results}.

As illustrated in Fig.~\ref{fig:7}, marked discrepancies in attack effectiveness are observed across the evaluated methods. Bbox-Att exhibits a relatively low ASR, indicating limited potency in compromising the detectors. Conversely, ColorFD yields a substantial increase in ASR while concurrently leading to a significant reduction in AP$_{50}$. Compared with white-box methods, ColorFD achieves higher ASR and lower AP$_{50}$ on YOLOv5u and Faster R-CNN. Meanwhile, it maintains comparable performance on YOLOv3u. Such results demonstrate that the proposed method possesses robust attack capabilities even within a black-box setting.

\begin{table*}[t]
	\centering
	\caption{Comparison of digital-space attack results obtained
		using ColorFD and other attack methods on three object detectors.
		Higher ASR and lower AP$_{50}$ indicate stronger attack performance.
		Bold values indicate the best result for each detector and metric.}
	\label{tab3:comparison_attack_results}
	
	\setlength{\tabcolsep}{5pt}
	\begin{tabular*}{\textwidth}
		{@{\extracolsep{\fill}}lcccccc}
		\toprule
		\multirow{2}{*}{Method}
		& \multicolumn{2}{c}{YOLOv3u}
		& \multicolumn{2}{c}{YOLOv5u}
		& \multicolumn{2}{c}{Faster R-CNN} \\
		\cmidrule(lr){2-3}
		\cmidrule(lr){4-5}
		\cmidrule(lr){6-7}
		& ASR (\%) & AP$_{50}$
		& ASR (\%) & AP$_{50}$
		& ASR (\%) & AP$_{50}$ \\
		\midrule
		Bbox-Att
		& 43.49 & 0.5621
		& 42.51 & 0.5482
		& 49.63 & 0.4967 \\
		
		AP-PA
		& 69.53 & \textbf{0.1722}
		& 58.97 & 0.3172
		& 89.43 & 0.0878 \\
		
		BADEI
		& 77.15 & 0.2249
		& 77.64 & 0.2182
		& 94.84 & 0.0499 \\
		
		ColorFD (unguided)
		& 71.50 & 0.2562
		& 70.76 & 0.2150
		& \textbf{95.09} & \textbf{0.0169} \\
		
		ColorFD (key-region)
		& \textbf{78.87} & 0.1888
		& \textbf{79.61} & \textbf{0.1453}
		& 88.21 & 0.0974 \\
		\bottomrule
	\end{tabular*}
	
%	\vspace{2pt}
%	\parbox{\textwidth}{\footnotesize
%		\textit{Note:} ASR denotes attack success rate.
%		AP$_{50}$ denotes average precision at an IoU threshold of 0.5.
%		Higher ASR and lower AP$_{50}$ indicate stronger attack performance.
%		Bold values indicate the best result for each detector and metric.
%	}
\end{table*}

The quantitative results in Table~\ref{tab3:comparison_attack_results} substantiate these observations. Specifically, ColorFD achieves superior ASR and lower AP$_{50}$ compared to Bbox-Att across all detectors, markedly outperforming the tested black-box patch method. On YOLOv5u, key-region-guided ColorFD yields an ASR of $79.61\%$, exceeding BADEI and AP-PA by $1.97$ and $20.64$ percentage points, respectively. Concurrently, it reduces AP$_{50}$ to $0.1453$, substantially lower than the values reported for BADEI ($0.2182$) and AP-PA ($0.3172$). Regarding Faster R-CNN, the unguided ColorFD reduces AP$_{50}$ to $0.0169$. This performance surpasses both BADEI ($0.0499$) and AP-PA ($0.0878$), indicating that the proposed method matches or even exceeds the efficacy of white-box attacks on specific models.

\begin{table*}[t]
	\centering
	\caption{Comparison of digital-space attack results obtained
		using ColorFD and other attack methods on dual-engine
		wing-mounted jet aircraft across three object detectors.
		Higher ASR and lower AP$_{50}$ indicate stronger attack performance.
		Bold values indicate the best result for each detector and metric.}
	\label{tab4:comparison_attack_common_results}
	
	\setlength{\tabcolsep}{5pt}
	\begin{tabular*}{\textwidth}
		{@{\extracolsep{\fill}}lcccccc}
		\toprule
		\multirow{2}{*}{Method}
		& \multicolumn{2}{c}{YOLOv3u}
		& \multicolumn{2}{c}{YOLOv5u}
		& \multicolumn{2}{c}{Faster R-CNN} \\
		\cmidrule(lr){2-3}
		\cmidrule(lr){4-5}
		\cmidrule(lr){6-7}
		& ASR (\%) & AP$_{50}$
		& ASR (\%) & AP$_{50}$
		& ASR (\%) & AP$_{50}$ \\
		\midrule
		Bbox-Att
		& 43.13 & 0.5655
		& 42.81 & 0.5505
		& 50.98 & 0.4833 \\
		
		AP-PA
		& 70.26 & \textbf{0.1458}
		& 61.11 & 0.2864
		& 93.14 & 0.0383 \\
		
		BADEI
		& \textbf{80.39} & 0.1950
		& 79.09 & 0.2024
		& 98.69 & 0.0118 \\
		
		ColorFD (unguided)
		& 72.22 & 0.2411
		& 71.57 & 0.2037
		& 97.39 & 0.0045 \\
		
		ColorFD (common-feature)
		& 77.12 & 0.1912
		& 75.16 & 0.1658
		& \textbf{99.35} & \textbf{0.0003} \\
		
		ColorFD (key-region)
		& 78.10 & 0.1949
		& \textbf{80.39} & \textbf{0.1459}
		& 87.91 & 0.0840 \\
		\bottomrule
	\end{tabular*}
	
%	\vspace{2pt}
%	\parbox{\textwidth}{\footnotesize
%		\textit{Note:} ASR denotes attack success rate.
%		AP$_{50}$ denotes average precision at an IoU threshold of 0.5.
%		Higher ASR and lower AP$_{50}$ indicate stronger attack performance.
%		Bold values indicate the best result for each detector and metric.
%	}
\end{table*}

Additional evaluations on the dual-engine aircraft subcategory are detailed in Table~\ref{tab4:comparison_attack_common_results}. On Faster R-CNN, common-feature guidance achieves an ASR of $99.35\%$, which is superior to BADEI ($98.69\%$). Furthermore, it reduces AP$_{50}$ to $0.0003$, representing the most effective result among all tested methods. On YOLOv5u, this strategy attains an ASR of $75.16\%$, which is significantly higher than that of AP-PA ($61.11\%$) and competitive with BADEI ($79.09\%$). These findings suggest that the incorporation of common-feature guidance enables effective cross-instance knowledge transfer. This characteristic is particularly beneficial for enhancing attack performance on two-stage detectors.

\begin{figure*}[!htb]
	\centering
	\includegraphics[width=2.0\columnwidth]{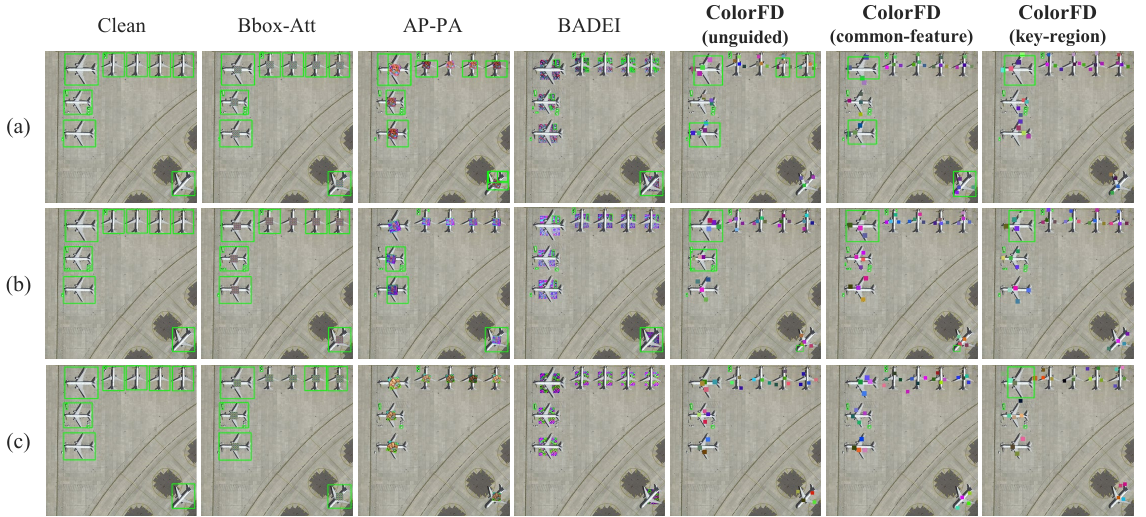}
	\caption{Detection results of different attack methods on three object detectors. ColorFD outperforms the tested black-box method and is comparable to white-box methods. Visual comparison of detection performance: (a) YOLOv3u; (b) YOLOv5u; (c) Faster R-CNN.}
	\label{fig:8}
\end{figure*}

Representative qualitative examples are presented in Fig.~\ref{fig:8}. ColorFD effectively suppresses detection results across all models. Conversely, Bbox-Att and AP-PA still produce residual detections with non-negligible confidence in complex scenarios. These visual observations align with the quantitative data and further validate the effectiveness of the proposed approach.

Overall, ColorFD consistently outperforms the tested black-box attack method. It achieves performance levels comparable to, or exceeding, state-of-the-art white-box attacks. Such results highlight the potency of the proposed method in black-box scenarios. They further demonstrate that the introduced guidance strategies provide substantial improvements in both optimization effectiveness and query efficiency.

\subsection{Physical-World Adversarial Attack}

Adversarial attack experiments are conducted within a physical environment to validate the feasibility of the proposed method in a controlled imaging setup. By utilizing a fixed camera position and viewing angle, the setup captures images of four alloy aircraft models. The inclusion of diverse aircraft, such as B777 and A380 variants, ensures the evaluation encompasses various aircraft categories and scales. Ultimately, this configuration serves to verify the cross-domain consistency of adversarial effectiveness.

\begin{figure}[!htb]
	\centering
	\includegraphics[width=0.85\columnwidth]{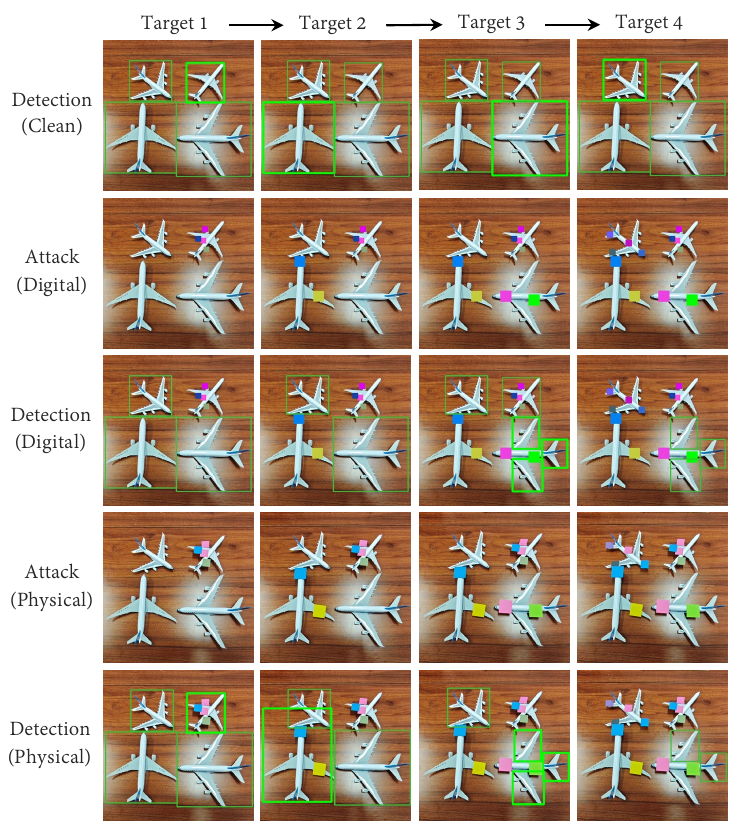}
	\caption{Comparison of digital and physical adversarial attack results for ColorFD. Each column corresponds to a different target aircraft, with the adversarial patch applied only to the designated aircraft. The patched aircraft is consistently missed by the detector in both digital and physical settings.}
	\label{fig:9}
\end{figure}

In an effort to approximate the digital pure-color patch design, physical patches are fabricated using colored paper sheets with 75 predefined colors. These patches are subsequently attached to target surfaces based on the optimized locations identified in the digital domain. YOLOv5u is employed as a representative detector to evaluate the resulting attack performance. Fig.~\ref{fig:9} illustrates the comparative results before and after the adversarial deployment.

As illustrated in Fig.~\ref{fig:9}, the colors of the physical patches deviate slightly from their digital counterparts due to material properties and varying imaging conditions. Furthermore, minor positional offsets occur because of target surface curvature and manual attachment inaccuracies. Despite these deviations, ColorFD effectively suppresses YOLOv5u detections for all evaluated aircraft targets under the tested imaging conditions. The physical results are consistent with the corresponding digital results, demonstrating the digital-to-physical transferability of the pure-color patches under the current controlled setup.

\section{Discussion}
\label{Sect:6}
\subsection{Hyperparameter Sensitivity Analysis}

The performance of ColorFD is influenced by key hyperparameters related to the Differential Evolution (DE) optimization process and patch configuration. Sensitivity analyses are conducted on the number of iterations, the population size multiplier (popmul), and the patch ratio to justify the selected configurations. These evaluations are performed on YOLOv5u and Faster R-CNN under the key-region guided setting, utilizing ASR and AP$_{50}$ as primary metrics.

\begin{figure}[!htbp]
	\centering
	\subfloat[]{%
		\includegraphics[width=0.83\columnwidth]{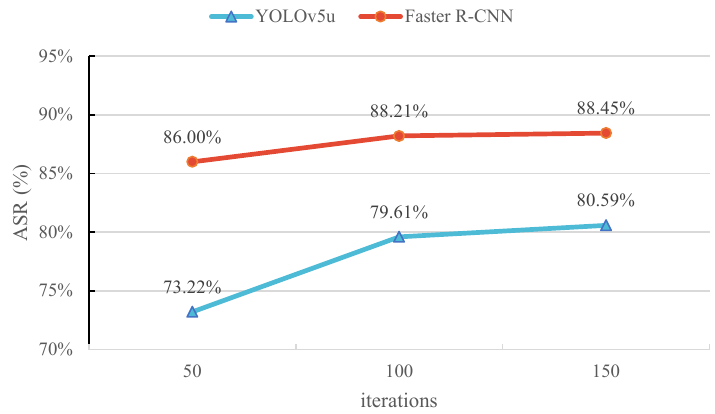}
		\label{fig:10.a}}\\[1mm]
	\subfloat[]{%
		\includegraphics[width=0.83\columnwidth]{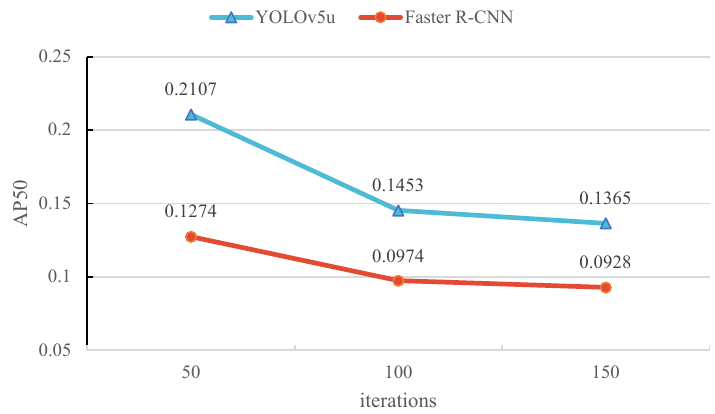}
		\label{fig:10.b}}
	\caption{Impact of the DE iteration number on attack performance. Both ASR and AP$_{50}$ show marked improvements from 50 to 100 iterations, whereas gains from 100 to 150 iterations are marginal.
		(a) ASR comparison.
		(b) AP$_{50}$ comparison.}
	\label{fig:10}
\end{figure}

\textbf{Iteration.}
The iteration determines the total number of population updates, thereby directly affecting the optimization depth of perturbations. Fig.~\ref{fig:10} illustrates the attack performance across different iteration settings (50, 100, and 150). A substantial improvement in ASR and a corresponding decline in AP$_{50}$ are observed on both detectors as the iteration increases from 50 to 100. This trend suggests that higher iterations facilitate the discovery of more potent adversarial perturbations. However, further increasing the count to 150 yields only marginal gains, indicating that the optimization process is approaching convergence. To balance attack effectiveness with computational cost, 100 iterations are adopted as a suitable trade-off.

\begin{figure}[!htb]
	\centering
	\subfloat[]{
		\includegraphics[width=0.83\columnwidth]{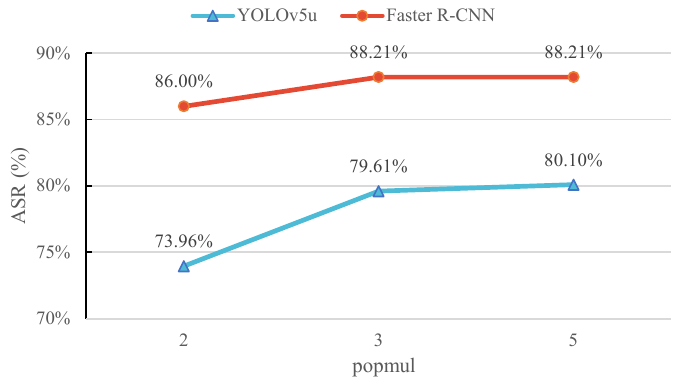}
		\label{fig:11.a}}\\[1mm]
	\subfloat[]{%
		\includegraphics[width=0.83\columnwidth]{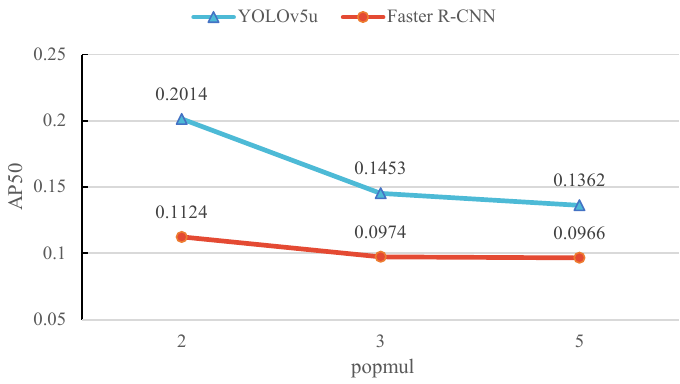}
		\label{fig:11.b}}
	\caption{Impact of the DE popmul value on attack performance. Notable improvements are observed from 2 to 3 population multipliers, while the gains from 3 to 5 remain marginal.
		(a) ASR comparison.
		(b) AP$_{50}$ comparison.}
	\label{fig:11}
\end{figure}

\textbf{Population Size Multiplier.}
The population size determines the number of candidate solutions explored per generation, which dictates the exploration capability of the search process. Fig.~\ref{fig:11} presents the results under various popmul settings (2, 3, and 5). Elevating popmul from 2 to 3 leads to enhanced ASR and reduced AP$_{50}$ on both detectors. Nevertheless, further increments to 5 result in diminishing returns, with only minimal performance gains observed. Consequently, the population size multiplier is set to 3.

\begin{figure}[!htb]
	\centering
	\subfloat[]{%
		\includegraphics[width=0.85\columnwidth]{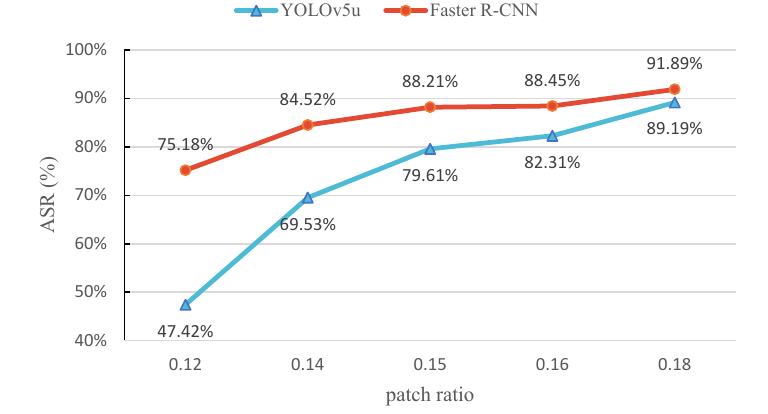}
		\label{fig:12.a}}\\[1mm]
	\subfloat[]{%
		\includegraphics[width=0.85\columnwidth]{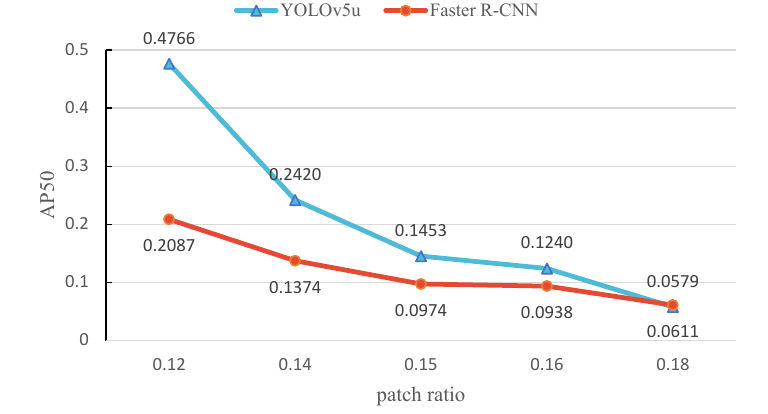}
		\label{fig:12.b}}
	\caption{Impact of the patch ratio value on attack performance. Notable improvements are observed from 0.12 to 0.15 patch ratio, while the gains from 0.15 to 0.18 remain marginal.
		(a) ASR comparison.
		(b) AP$_{50}$ comparison.}
	\label{fig:12}
\end{figure}

\textbf{Patch Ratio.}
The patch ratio denotes the relative size of each pure-color patch with respect to the target bounding box. A smaller patch ratio results in less conspicuous perturbations. Fig.~\ref{fig:12} presents the results under various patch ratio settings (0.12, 0.14, 0.15, 0.16, and 0.18). For both YOLOv5u and Faster R-CNN, ASR increases as the patch ratio increases. However, the improvement from 0.15 to 0.18 is considerably smaller than that from 0.12 to 0.15, indicating limited marginal gains beyond 0.15. A similar trend is observed for AP$_{50}$, whose reduction becomes less pronounced when the patch ratio exceeds 0.15. To maintain strong attack performance while minimizing perturbation visibility, the patch ratio is set to 0.15.

\subsection{Limitations}

Despite achieving strong performance in both digital and physical domains, the proposed method presents several limitations.
First, the stochastic nature of the DE-based black-box optimization leads to performance variability across different runs. For certain samples, multiple optimization attempts are required to achieve satisfactory results.
Second, the effectiveness of the key-region localization strategy is contingent upon the underlying detection framework. While successful on one-stage detectors, a reduced efficacy is noted on two-stage models such as Faster R-CNN. This discrepancy is primarily attributed to the inaccessibility of intermediate RPN information in black-box settings. Consequently, the key-region localization strategy struggles to capture accurate region-specific responses. 
Finally, the pure-color patch design remains sensitive to environmental variations. Attack performance is highly dependent on stable imaging conditions. Significant shifts in viewpoint, scale, or camera position can disrupt the alignment between patches and critical target regions, thereby degrading attack performance.

\section{Conclusion}
\label{Sect:7}
This paper presents ColorFD, a black-box physical attack based on multiple pure-color patches for remote sensing object detectors. To address the black-box settings, the patch positions and color parameters are jointly optimized using DE with access only to the detection results returned by the target detector. A target-wise fitness and selection mechanism is introduced to improve multi-target optimization. The target-level fitness evaluates the attack state of each target separately. The selection mechanism preserves target-specific improvements while maintaining the overall fitness. Furthermore, key-region guidance identifies instance-specific sensitive regions through finite-difference color probing. Common-feature guidance further provides category-level spatial priors and reduces repeated localization.

Experiments on the DIOR-based aircraft subset demonstrate that ColorFD effectively degrades the detection performance of YOLOv3u, YOLOv5u, and Faster R-CNN. Compared with Bbox-Att, ColorFD achieves substantial improvements in both ASR and AP$_{50}$. Its attack performance remains comparable to strong white-box baselines like AP-PA and BADEI. Physical-world experiments further validate the feasibility of digital-to-physical transfer under the tested imaging conditions. These results demonstrate the effectiveness of the pure-color patch design, target-wise optimization, and finite-difference guided constraint strategies. Nevertheless, the stochastic nature of DE, architectural constraints of the detectors, and environmental variations remain limitations. Future work will focus on improving optimization stability and the visual stealthiness of adversarial patches.

\bibliographystyle{IEEEtran}
\bibliography{references}

\vspace{-30pt}
\begin{IEEEbiography}[{\includegraphics[width=1in,height=1.25in,clip,keepaspectratio]{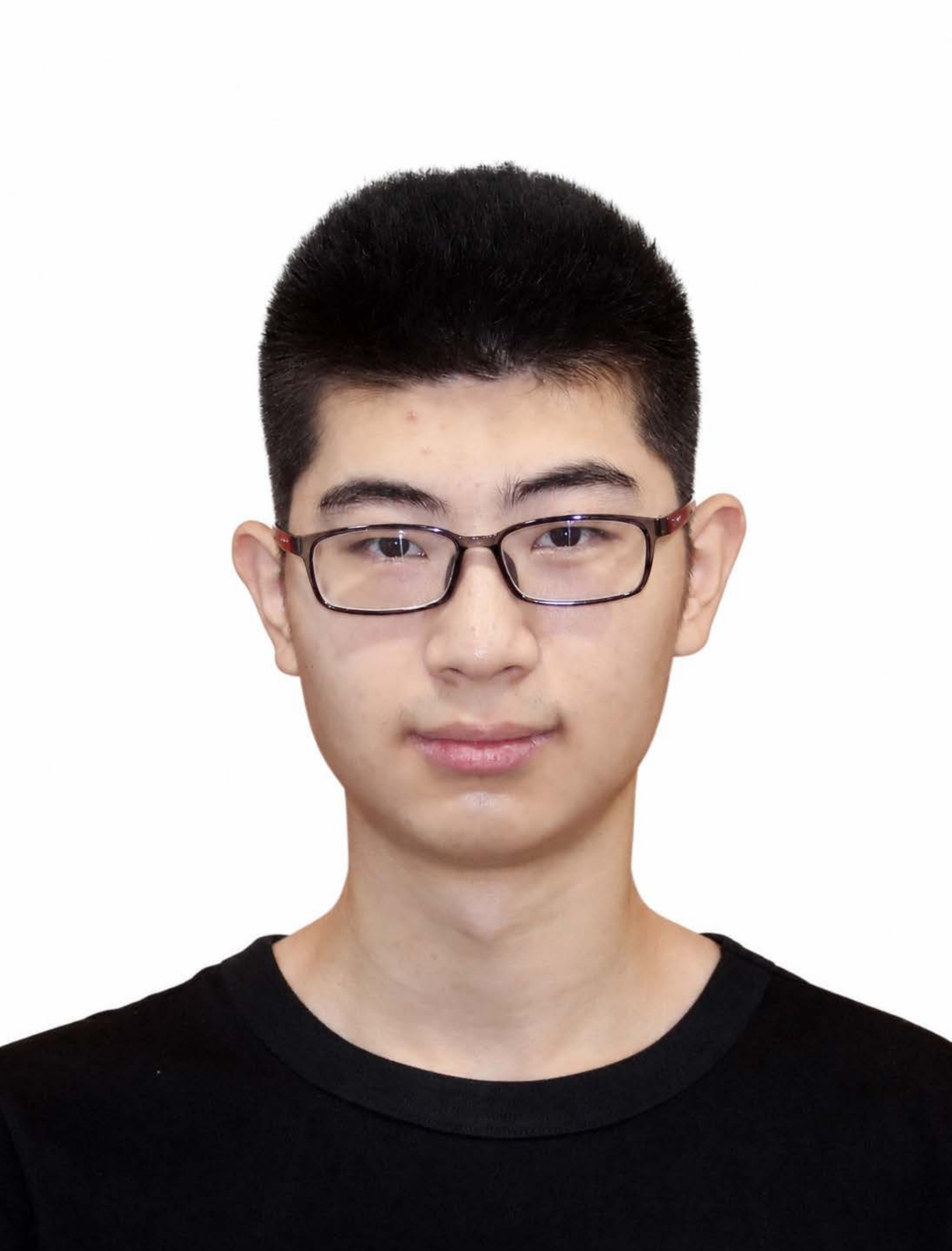}}]{Tiannuo Guo}
	received the B.S. degree in computer science and technology from China University of Mining and Technology in 2024. He is currently pursuing the M.S. degree with the College of Information Science and Technology, Beijing University of Chemical Technology, Beijing, China.
	
	His research interests include computer vision, adversarial attacks, and generative models.
\end{IEEEbiography}
\vspace{-30pt}
\begin{IEEEbiography}[{\includegraphics[width=1in,height=1.25in,clip,keepaspectratio]{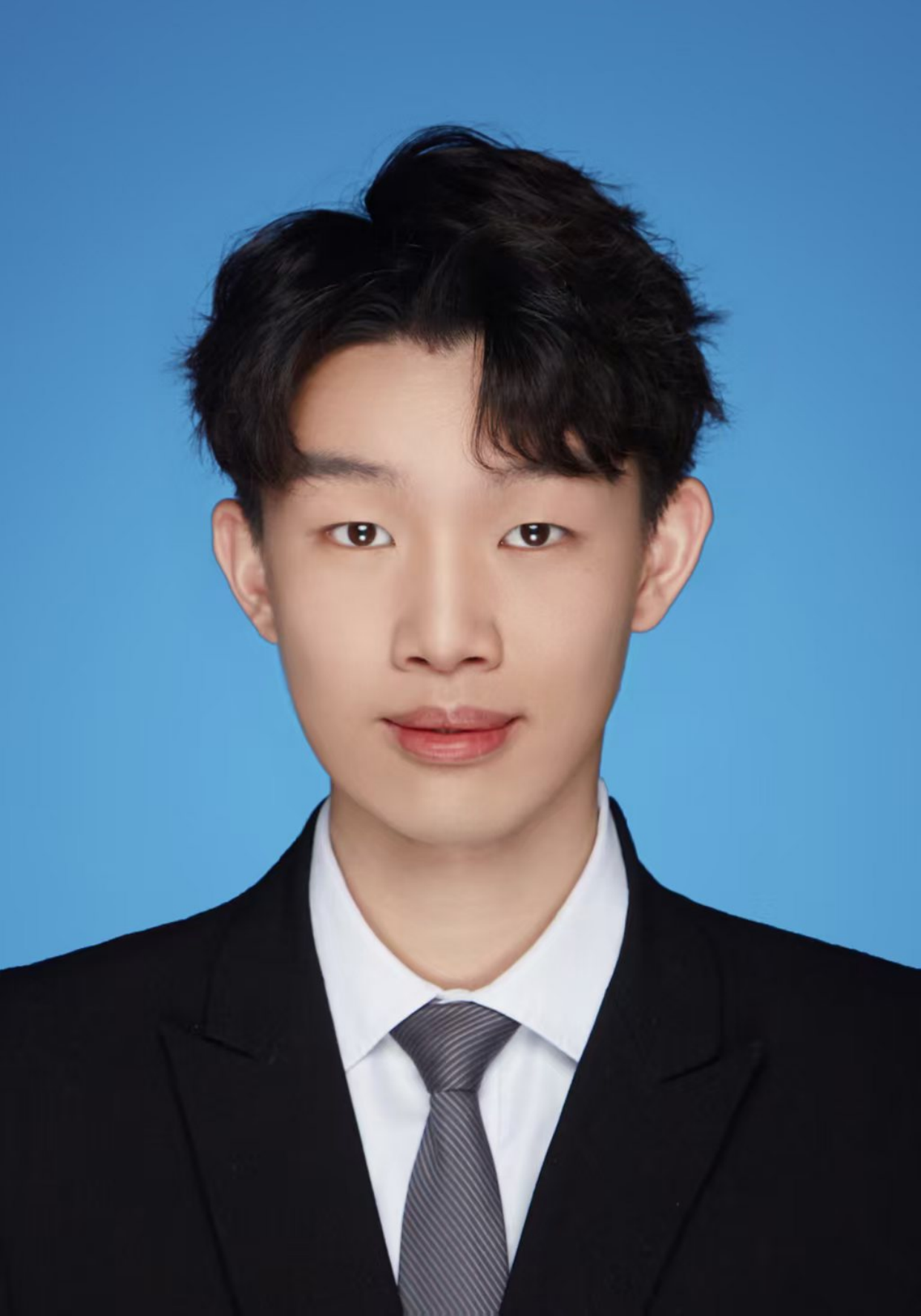}}]{Guhang Qiu}
	received the B.S. degree in artificial intelligence from Beijing University of Chemical Technology in 2024. He is currently pursuing the M.S. degree with the College of Information Science and Technology, Beijing University of Chemical Technology, Beijing, China.
	
	His research interests include remote sensing image processing, deep learning, machine learning and computer vision applications.
\end{IEEEbiography}
\vspace{-30pt}
\begin{IEEEbiography}[{\includegraphics[width=1in,height=1.25in,clip,keepaspectratio]{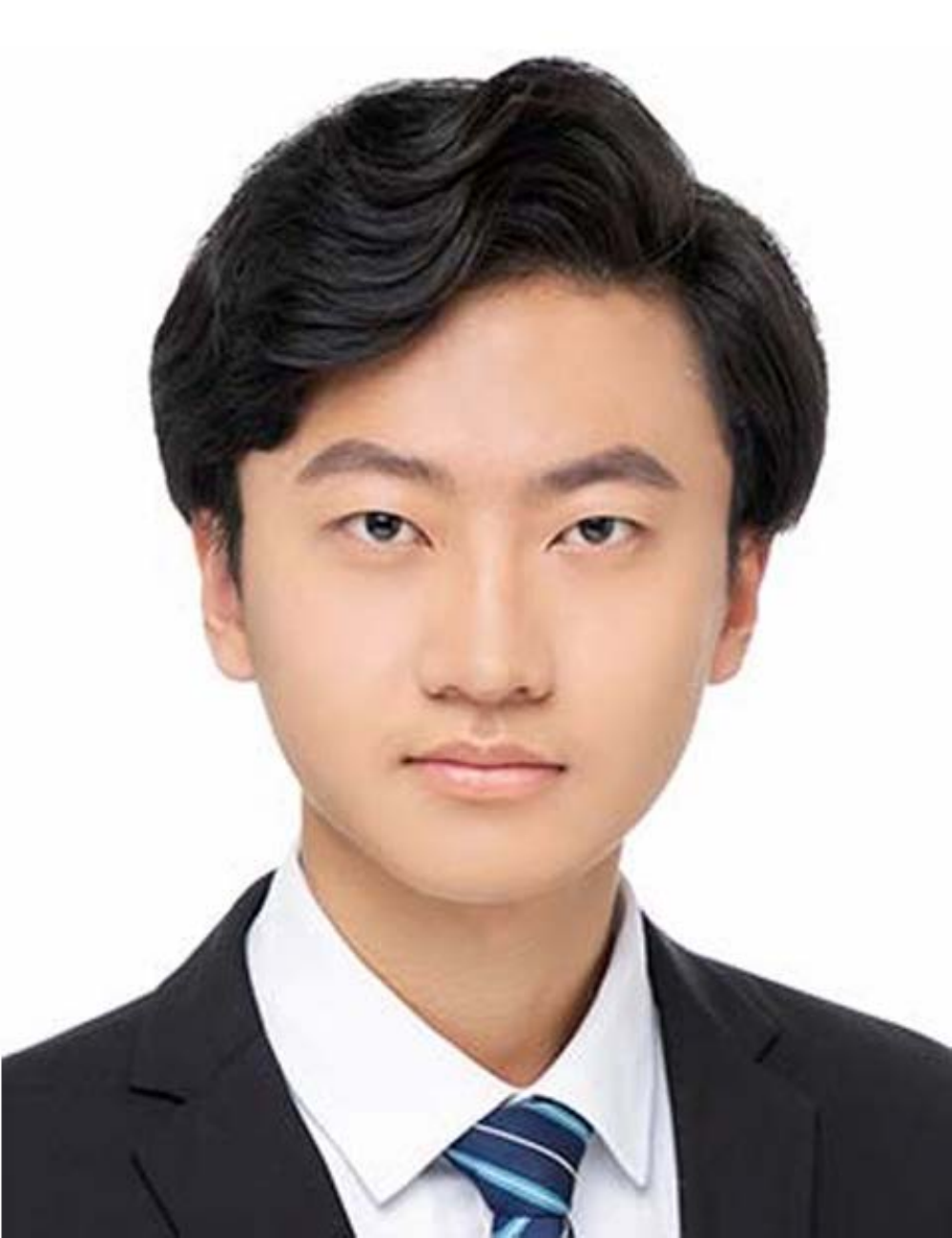}}]{Yuzhen Xie}
	received the M.S. degree in information science and technology in 2023 from the School of Information Science and Technology, Beijing University of Chemical Technology, Beijing, China, where he is currently working toward the Ph.D. degree in control science and engineering with the School of Information Science and Technology.

	His major research interests include target recognition in synthetic aperture radar (SAR) and polarimetric SAR imagery, deep learning, and its applications in remote sensing image analysis.
\end{IEEEbiography}
\vspace{-30pt}
\begin{IEEEbiography}[{\includegraphics[width=1in,height=1.25in,clip,keepaspectratio]{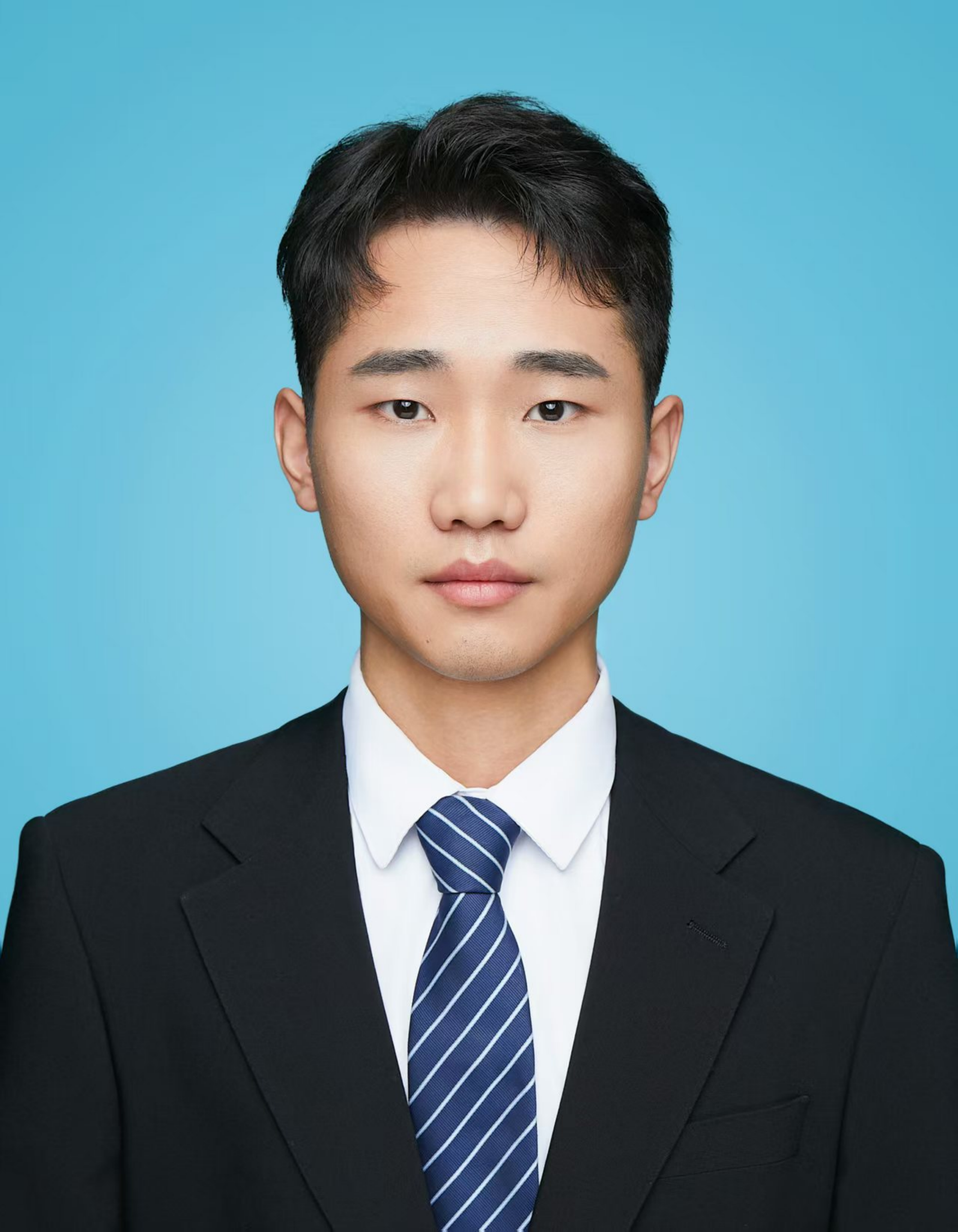}}]{Rui Feng}
	received the M.Eng. degree from Beijing University of Chemical Technology, Beijing, China, in 2025, where he is currently pursuing the Ph.D. degree with the College of Information Science and Technology.
	
	His research interests include SAR target image generation, PolSAR change detection, and related intelligent interpretation methods.
\end{IEEEbiography}
\vspace{-30pt}
\begin{IEEEbiography}[{\includegraphics[width=1in,height=1.25in,clip,keepaspectratio]{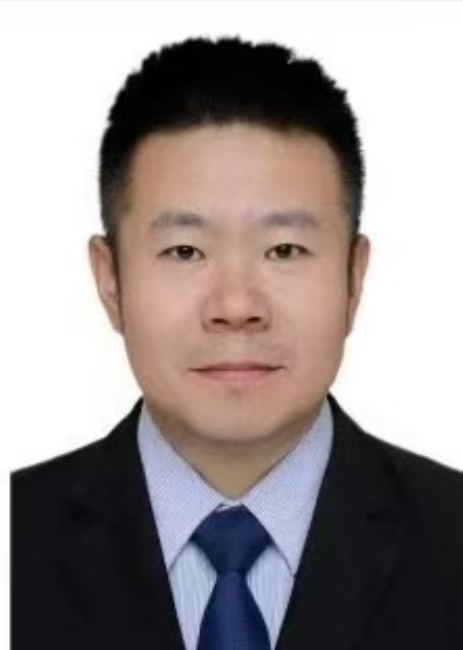}}]{Ligang Li}
	received the Ph.D. degree in optical engineering from the Chinese Academy of Sciences, Beijing, China, in 2006. He is currently a Professor with Beijing University of Chemical Technology, Beijing, China.
	
	His research interests include optical imaging simulation system technology, spatial information processing, and other research topics.
\end{IEEEbiography}
\vspace{-30pt}
\begin{IEEEbiography}[{\includegraphics[width=1in,height=1.25in,clip,keepaspectratio]{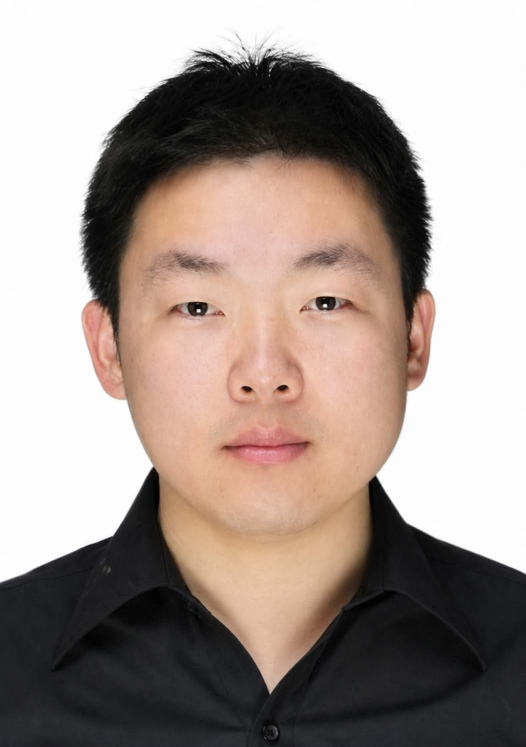}}]{Deliang Xiang}
	(Member, IEEE) received the B.S. degree in remote sensing science and technology from Wuhan University, Wuhan, China, in 2010, the M.S. degree in photogrammetry and remote sensing from the National University of Defense Technology, Changsha, China, in 2012, and the Ph.D. degree in geoinformatics from the KTH Royal Institute of Technology, Stockholm, Sweden, in 2016. Since 2020, he has been a Full Professor with the College of Information Science and Technology, Beijing University of Chemical Technology, Beijing, China.
	
	His research interests include urban remote sensing, synthetic aperture radar (SAR)/polarimetric SAR image processing, artificial intelligence, and pattern recognition.
\end{IEEEbiography}

\end{document}